\documentclass[
]{ceurart}

\usepackage{listings}
\usepackage{acronym}
\acrodef{BioNLP}[BioNLP]{Biomedical Natural Language Processing}
\acrodef{C-RE}[C-RE]{Concept-level RE}
\acrodef{CI}[CI]{Confidence Interval}
\acrodef{CS}[CS]{Cluster Sampling}
\acrodef{IAA}[IAA]{Inter-Annotator Agreement}
\acrodef{IE}[IE]{Information Extraction}
\acrodef{GSC}[GSC]{Global Surface-form Clustering}
\acrodef{KBC}[KBC]{Knowledge Base Construction}
\acrodef{KG}[KG]{Knowledge Graph}
\acrodef{KOS}[KOS]{Knowledge Organization System}
\acrodef{LLM}[LLM]{Large Language Model}
\acrodef{M-RE}[M-RE]{Mention-level RE}
\acrodef{MoE}[MoE]{Margin of Error}
\acrodef{NER}[NER]{Named Entity Recognition}
\acrodef{NEL}[NEL]{Named Entity Linking}
\acrodef{NLP}[NLP]{Natural Language Processing}
\acrodef{OIE}[OIE]{Open Information Extraction}
\acrodef{PPS}[PPS]{Probability Proportional to Size}
\acrodef{PPS-WR}[PPS-WR]{Probability Proportional to Size With Replacement}
\acrodef{PPS-WOR}[PPS-WOR]{Probability Proportional to Size Without Replacement}
\acrodef{RCS}[RCS]{Random Cluster Sampling}
\acrodef{RE}[RE]{Relation Extraction}
\acrodef{SOTA}[SOTA]{State of the Art}
\acrodef{SRS}[SRS]{Simple Random Sampling}
\acrodef{SRS-WR}[SRS-WR]{Simple Random Sampling With Replacement}
\acrodef{SRS-WOR}[SRS-WOR]{Simple Random Sampling Without Replacement}
\acrodef{SS}[SS]{Stratified Sampling}
\acrodef{STWCS}[STWCS]{Stratified Two-stage Weighted Cluster Sampling}
\acrodef{TWCS}[TWCS]{Two-stage Weighted Cluster Sampling}
\acrodef{URI}[URI]{Uniform Resource Identifier}
\acrodef{WCS}[WCS]{Weighted Cluster Sampling}

\usepackage{comment}

\begin{document}

\copyrightyear{2025}
\copyrightclause{Copyright for this paper by its authors.
  Use permitted under Creative Commons License Attribution 4.0
  International (CC BY 4.0).}

\conference{Submitted to IRCDL'26: 22nd Conference on Information and Research Science Connecting to Digital and Library Science,
  February 19--20, 2026, Modena, Italy}

\title{Efficient and Reliable Estimation of Named Entity Linking Quality: A Case Study on GutBrainIE}

%\tnotetext[1]{You can use this document as the template for preparing your
%  publication. We recommend using the latest version of the ceurart style.}

\author[1]{Marco Martinelli}[%
orcid=0009-0001-1596-8642,
email=martinell2@dei.unipd.it,
url=https://www.dei.unipd.it/~martinell2/,
]
\cormark[1]

\author[1]{Stefano Marchesin}[%
orcid=0000-0003-0362-5893,
email=stefano.marchesin@unipd.it,
url=https://www.dei.unipd.it/~marches1/,
]

\author[1]{Gianmaria Silvello}[%
orcid=0000-0003-4970-4554,
email=gianmaria.silvello@unipd.it,
url=https://www.dei.unipd.it/~silvello/,
]

\address[1]{Department of Information Engineering, University of Padua, Italy}
\cortext[1]{Corresponding author.}

\begin{abstract}
Named Entity Linking (NEL) is a core component of biomedical Information Extraction (IE) pipelines, yet assessing its quality at scale is challenging due to the high cost of expert annotations and the large size of corpora. 
In this paper, we present a sampling-based framework to estimate the NEL accuracy of large-scale IE corpora under statistical guarantees and constrained annotation budgets. %; we focus on the \textsc{GutBrainIE} collection openly released in fall 2025.   
We frame \ac{NEL} accuracy estimation as a constrained optimization problem, where the objective is to minimize expected annotation cost subject to a target Margin of Error (MoE) for the corpus-level accuracy estimate. 
Building on recent works on knowledge graph accuracy estimation, we adapt Stratified Two-Stage Cluster Sampling (STWCS) to the NEL setting, defining label-based strata and global surface-form clusters in a way that is independent of NEL annotations. 
Applied to 11,184 NEL annotations in \textsc{GutBrainIE} -- a new biomedical corpus openly released in fall 2025 -- our framework reaches a MoE $\leq 0.05$ by manually annotating only 2,749 triples (24.6\%), leading to an overall accuracy estimate of $0.915 \pm 0.0473$. A time-based cost model and simulations against a Simple Random Sampling (SRS) baseline show that our design reduces expert annotation time by about 29\% at fixed sample size. The framework is generic and can be applied to other NEL benchmarks and IE pipelines that require scalable and statistically robust accuracy assessment.
\end{abstract}


\maketitle

\section{Introduction}

\ac{NEL} is a key component in \ac{IE} pipelines, especially in biomedical domains where downstream tasks such as \ac{RE} and \ac{KBC} strictly depend on the correctness of entity-concept mappings \cite{shen2014entity,al2020named,french2023overview}. 
These tasks play a crucial role in building reliable automatic \ac{IE} systems that can support clinicians and researchers in processing large volumes of unstructured text, for instance when conducting biomedical literature reviews or extracting structured information from triage records and other clinical documents \cite{jonnalagadda2015automating,wang2018clinical,stewart2023applications}.  
However, assessing the quality of \ac{NEL} annotations at scale is challenging for two reasons: expert annotations are expensive, and manually revising all linked mentions one by one is often infeasible \cite{
qi2022evaluating,marchesin2024utility}.

In this work, we tackle the challenge of estimating the accuracy of \ac{NEL} in large-scale \ac{IE} corpora, with a focus on scalability and cost-efficiency.  %Given that most of these collections contain a large number of linkages, manually validating each of them would be really costly and often practically infeasible.  
To overcome this, we introduce an iterative, sampling-based framework that selectively draws triples for manual verification and leverages those verifications to estimate corpus-level accuracy with formal statistical guarantees. The core idea is to repeatedly sample a small, representative subset of triples (mention-concept linkages in the present context), obtain manual judgments on them, and use these results to refine the overall accuracy estimate. This process continues until the estimate falls within a predefined margin of error. The proposed estimation framework rests on two foundational elements: establishing a statistical \ac{CI} and selecting an appropriate sampling strategy to meet that requirement efficiently. The \ac{CI} specifies the acceptable level of uncertainty in our accuracy estimate, while the sampling strategy determines how triples are selected for manual verification to achieve this confidence level at minimal cost.

We validate this framework through a case study on the \textsc{GutBrainIE} collection \cite{martinelli2025overview,nentidis2025overview}, a large-scale corpus on the gut-brain axis \cite{carabotti2015gut,appleton2018gut}, featuring over 11'000 concept-level linkages. 

\begin{comment}
In this work, we address the problem of estimating the \ac{NEL} accuracy of the \textsc{GutBrainIE} collection, a large-scale corpus on the gut-brain axis \cite{carabotti2015gut,appleton2018gut}, while minimizing expert effort and providing statistical guarantees. 
Given the large number of linkages included in the \textsc{GutBrainIE} collection, manually validating each of these  
%Instead of manually reviewing all linkages, we design an iterative sampling-based framework that draws targeted facts to be manually verified and then estimates corpus-level accuracy with \acp{CI}. %and controlled \ac{MoE}.
\end{comment}

Our main contributions are:
\begin{itemize}
    \item We formalize \ac{NEL} accuracy estimation as a constrained optimization problem, where the goal is to minimize annotation cost under a target \ac{CI} constraint.
    \item We adapt \ac{STWCS}, a \ac{SOTA} sampling strategy developed for \ac{KG} accuracy estimation \cite{gao2019efficient}, to the \ac{NEL} task. %Our design uses label-based strata and global surface-form clusters, remaining \ac{NEL}-independent.
    \item We evaluate the framework on \textsc{GutBrainIE}, showing that annotating less than 25\% of all triples is enough to obtain a corpus-level accuracy estimate with MoE $\leq 0.05$, and that our \ac{STWCS}-based framework reduces total annotation time by about 29\% compared to typical adopted settings such as \ac{SRS} with a fixed sample size.
\end{itemize}

The remainder of the paper is organized as follows: Section \ref{sec:Background and Task Setting} presents the background on accuracy estimation for \acp{KG} and introduces the \textsc{GutBrainIE} collection; Section \ref{sec:accuracyEstimationForFactCollections} formalizes the employed \ac{NEL} accuracy estimation framework, describing the definition of strata and clusters, the resulting estimator, and the considered cost model; Section \ref{sec:Results} reports the results obtained on \textsc{GutBrainIE}, including corpus-level and strata-level accuracy estimates along with an efficiency analysis comparing our framework against a \ac{SRS} baseline; finally, Section \ref{sec:Conclusion} draw conclusions and discusses directions for future work.   

\section{Background and Task Setting}
\label{sec:Background and Task Setting}

\subsection{Sampling and Accuracy Estimation in \acp{KG}}
\label{sec:backgroundAccuracyEstimation}
In sampling-based frameworks for \ac{KG} accuracy estimation, the goal is to approximate the accuracy of the \ac{KG} (proportion of correct triples over the total) by manually reviewing only a small subset of triples drawn with a certain sampling design \cite{gao2019efficient,marchesin2024efficient,marchesin2024utility}.
The sampling design determines the statistical properties of the resulting estimator (e.g., variance, bias), the amount of human effort required, and, therefore, the overall annotation cost.

In this work, with \textit{sampling} we mean drawing a subset of triples $T_S$ from the full collection $T$ and annotating only this subset to estimate the global accuracy $\mu(T)$.
$S$ is defined as the employed \textit{sampling strategy}. We consider $S$ as based on some sort of random sampling. Thus, the resulting estimator $\hat{\mu}(T_S)$ also exhibits some degree of randomness. 

Following prior work on \ac{KG} accuracy estimation \cite{gao2019efficient, marchesin2024efficient, marchesin2024utility}, we associate an estimator $\hat{\mu}$ with a $(1-\alpha)$ \ac{CI}, that is, an interval which contains the true accuracy $\mu(T)$ with probability approximately $(1-\alpha)$. Namely, the true accuracy $\mu(T)$ is bounded in: 
\[
[\hat{\mu}-\frac{\text{CI}}{2}\;,\,\hat{\mu}+\frac{\text{CI}}{2}]
\]
By defining the \textit{\ac{MoE}} as the half-width of the \ac{CI}, we can rewrite the bounds for $\mu(T)$ as 
\[
[\hat{\mu}-\text{MoE}\;,\,\hat{\mu}+\text{MoE}]
\]
Since the \ac{MoE} shrinks as the sample size $|T_S\,|$ grows, the goal is to design $S$ to reach the target $\text{MoE} \leq \varepsilon$ while minimizing the number of manually annotated triples.

A straightforward choice for $S$ is \ac{SRS}, in which triples are selected uniformly at random from $T$ without replacement. Under \ac{SRS}, the sample mean provides an unbiased estimator of $\mu(T)$, and the corresponding \ac{CI} can be derived using standard normal approximations \cite{gao2019efficient}. However, \ac{SRS} inherently scatters sampled triples across many unrelated documents, requiring annotators to repeatedly reconstruct the surrounding context for each triple. Because \acp{KG} are often large and highly heterogeneous -- even within specialized domains -- \ac{SRS} may, for example, present an annotator with a triple about an actor immediately followed by one about a geographic location.  This contextual fragmentation creates inefficiency in both annotation time and cognitive burden.
Indeed, as shown in works on \ac{KG} accuracy estimation, sampling independent triples in most cases leads to increased annotation costs \cite{gao2019efficient, marchesin2024efficient}.

To address this, \citet{gao2019efficient} propose \textit{\ac{CS}}.
The idea is to group triples that share the same contextual feature (e.g., the same subject entity) into \textit{clusters}, and then sample clusters rather than individual triples for annotation.
The rationale is that, once an annotator revises a certain triple, annotating additional triples drawn from the same cluster is more efficient than annotating triples from different clusters.
Within this framework, two main cluster sampling designs are defined: \ac{RCS}, which samples clusters with a uniform random probability, and \ac{WCS}, which instead samples clusters with \ac{PPS}, i.e., proportional to the number of triples contained in each cluster.
Both designs yield unbiased estimators of global accuracy \cite{gao2019efficient}.

\ac{CS} designs still present a limitation: annotating every triple in selected clusters may be expensive, especially if considered clusters are large.
To address this, \citet{gao2019efficient} introduce \acf{TWCS}. In the first stage, clusters are sampled with \ac{PPS} (as in \ac{WCS}); in the second stage, up to $m$ triples are drawn without replacement from each sampled cluster and annotated.
The resulting estimator $\hat{\mu}$ averages the within-cluster accuracies of sampled clusters and is proven to be unbiased for the overall accuracy $\mu(T)$.
Intuitively, \ac{TWCS} reduces annotation cost by placing an upper bound $m$ on the number of triples annotated per cluster, while still leveraging the fact that sampling multiple triples within the same cluster is cheaper than sampling independent triples \cite{gao2019efficient}.

\ac{TWCS} efficiency can be further improved by introducing \textit{stratification}, i.e., partitioning the set of triples $T$ into non-overlapping \textit{strata} that are internally more homogeneous (e.g., in terms of expected accuracy, utility, or domain knowledge) than the entire set $T$ as a whole \cite{marchesin2024efficient, marchesin2024utility}.
Stratification can be applied before or after clustering without any particular loss of generality.
By merging stratification with \ac{TWCS}, we obtain \acf{STWCS}, a three-stage sampling design in which the first step is to sample a strata with \ac{PPS-WR}, considering as size the number of triples contained in the strata. The last two steps apply \ac{TWCS} as described above to the selected stratum. 
The rationale underlying \ac{STWCS} is that, by building strata, triples with similar accuracy are grouped together; thus, variability within each stratum is lower than in the population as a whole. As a result, the estimator needs fewer annotated triples to reach the same \ac{MoE}, and sampling can be focused on strata that have the most impact in reducing \ac{CI} width, instead of being uniformly spread over the entire population, as in standard (non-stratified) \ac{TWCS} \cite{marchesin2024utility}.  

\subsection{Named Entity Linking for \textsc{GutBrainIE}}
\label{sec:nel_gutbrainie}

\textsc{GutBrainIE} is a large-scale biomedical corpus of PubMed abstracts focused on the gut-brain axis, which includes manual and automatic annotations for entities, concept-level links, and relations \cite{martinelli2025overview,nentidis2025overview}. 
In recent years, the number of publications registered on PubMed on the gut-brain axis has been continuously increasing, with the yearly number of articles surpassing 2,000 in 2025.\footnote{The value was obtained with the query \texttt{(gut brain axis[Title/Abstract]) OR (gut brain axis[Title/Abstract])} on PubMed.}, making it increasingly difficult for field experts to stay up to date and extract relevant findings from this rapidly expanding body of literature. 
At the same time, gut-brain articles are characterized by complex, highly domain-specific terminology and semantics, which limit the effectiveness of generic biomedical \ac{IE} models and hinder transfer learning from other biomedical corpora \cite{jafari2025application}. 
\textsc{GutBrainIE} is designed to foster the development and evaluation of gut–brain-specific \ac{IE} systems that can support experts in automatically processing this literature \cite{martinelli2025overview},

\textsc{GutBrainIE} covers 13 entity types, including widely used biomedical categories (e.g., \textit{anatomical location}, \textit{bacteria}, \textit{drug}) and entities specific to the gut-brain axis (e.g., \textit{microbiome}, \textit{dietary supplement}). To account for the frequent occurrence of experimental scenarios in documents, the corpus also includes specific categories related to medical experiments (e.g., \textit{biomedical technique} and \textit{statistical technique}).

On the relation side, \textsc{GutBrainIE} features 17 relation predicates, many of which are overloaded and can connect different combinations of entity types depending on context. This many-to-many design originates 55 distinct relation triples. 
For \ac{NEL}, entity mentions are linked to concepts drawn from 6 standardized biomedical vocabularies, plus a custom ontology for unmatched mentions.
By ``\textit{linking to a concept}'' we mean assigning to the entity mention a \ac{URI} that resolves to the corresponding concept in a reference resource.

To ensure consistency across the corpus, \textsc{GutBrainIE} defines, for each entity type, a priority order over reference resources so that when the same concept is available in multiple resources, the selected concept is always drawn from the highest-priority one.
The full list of entity types and their associated resources is reported in Table \ref{tab:links_definitions}.

\begin{table}[!htb]
\centering
\caption{Biomedical resources used for concept-level linking of entity mentions. For each entity label, the table lists the reference resources ordered accordingly to the priority considered in \ac{NEL} annotations. \textit{GBIE} indicates our custom-defined vocabulary.}
\label{tab:links_definitions}
\resizebox{0.85\textwidth}{!}{%
\begin{tabular}{ll}
\hline
\textbf{Entity Label} & \textbf{Linked Vocabularies}                                        \\ \hline
Anatomical Location   & UMLS, NCIT, GBIE                                       \\ \hline
Animal                & UMLS, NCIT, NCBITaxon, GBIE                         \\ \hline
Bacteria              & UMLS,  NCIT, NCBITaxon, MESH, OMIT, GBIE \\ \hline
Biomedical Technique \qquad\qquad & UMLS, NCIT, OMIT, NCBITaxon, GBIE            \\ \hline
Chemical              & UMLS, NCIT, CHEBI, OMIT, GBIE              \\ \hline
Dietary Supplement & NCIT, UMLS, CHEBI, NCBITaxon, OMIT, MESH, GBIE \\ \hline
DDF                   & UMLS, NCIT, OMIT, NCBITaxon, GBIE         \\ \hline
Drug                  &  UMLS, NCIT, CHEBI, OMIT, NCBITaxon, GBIE  \\ \hline
Food                  & UMLS, NCIT, GBIE                                      \\ \hline
Gene                  & UMLS, NCIT, OMIT, CHEBI, GBIE                 \\ \hline
Human                 & UMLS, MESH, GBIE                                      \\ \hline
Microbiome            & UMLS, NCIT, NCBITaxon, GBIE                          \\ \hline
Statistical Technique & UMLS, NCIT, GBIE                                      \\ \hline
\end{tabular}%
}
\end{table}

The \textsc{GutBrainIE} collection includes a total of 1'647 documents and is organized into four folds that reflect annotation quality and annotator expertise. Platinum and Gold folds contain annotations curated by domain experts (399 documents), Silver is annotated by trained laypersons (499 documents), while Bronze is automatically annotated without subsequent manual revision (749 documents).
Considering manually annotated documents only, these present a high density of annotations, featuring, on average, 29.46 entities and 15.47 relations per document.

\ac{NEL} annotations are automatically generated by a linking system that combines heuristic normalization rules with lexical and similarity matching against the reference resources. 
For that reason, \ac{NEL} annotations are only for the expert-curated folds (Platinum and Gold), since layperson and automatic annotations lack the precision required for reliable automatic concept mapping.
In total, this pipeline produces more than 11,000 concept-level links. Manually revising all of them would be extremely costly and time-consuming. Therefore, we propose a statistically reliable framework to estimate overall \ac{NEL} accuracy while limiting manual revision costs and supporting targeted manual corrections instead of comprehensive re-annotation.

\section{Accuracy Estimation for \ac{NEL} Annotations} \label{sec:accuracyEstimationForFactCollections}
We follow previous work on sampling-based accuracy estimation for \acp{KG} (c.f., Section \ref{sec:backgroundAccuracyEstimation}), where the goal is to estimate the overall accuracy of \ac{KG} triples while minimizing the number of manual annotations and providing statistical guarantees about the resulting estimates. 
Without loss of generality, we can view each \ac{NEL} annotation in our setting as a triple in a \ac{KG}, where the mention (text span + entity label) acts as subject, the concept \ac{URI} as object, and a generic \textit{hasConcept} predicate connects them.
Therefore, we can directly apply the same methodologies used for \ac{KG} accuracy estimation to \ac{NEL} annotations.

Let $T$ be the set of all such triples. We define a correctness indicator function:
\[
\mathbb{1}_T\,(t) \in \{0,1\},
\]
where $\mathbb{1}_T\,(t) = 1$ if the triple is judged correct by an expert annotator, $0$ otherwise. A triple is considered incorrect if the entity mention is linked to a wrong concept or if the assigned concept is overly generic, i.e., a more specific and appropriate one exists in the reference resources. 
Our goal is to estimate the mean correctness of all such triples:
\begin{equation}
    \mu(T) = \frac{1}{|T|} \sum_{t \in T} \mathbb{1}_T\,(t),
\end{equation}
using manual annotations collected only for a subset $T_S$ of $T$.
Specifically, a sampling-based evaluation framework replaces $\mu(T)$ with an estimator $\hat{\mu}$ computed over a sample $T_S \subset T$ drawn according to some sampling strategy $S$. To be meaningful, the estimator should be (approximately) unbiased, i.e., $E[\hat{\mu}] = \mu(T)$, and accompanied by a $(1-\alpha)$ \ac{CI} with controlled \ac{MoE}, defined as half the \ac{CI} width. 
Unbiasedness is crucial because it guarantees that, on average, the accuracy estimated from the sampled triples $T_S$ reflects the true accuracy $\mu(T)$ of the entire set $T$. Moreover, this makes the associated \acp{CI} reliable in quantifying how well the estimated accuracy on $T_S$ reflects the corpus-level accuracy $\mu_T$.

\subsection{Sampling Design}
\label{sec:samplingDesign}
In this work, we adopt \ac{STWCS} as our sampling design and adapt it to \ac{NEL} triples of the \textsc{GutBrainIE} collection.
To do that, we need to define: 
(i) \textit{strata}, which partition triples into non-overlapping groups, typically to reduce variance and incorporate domain knowledge, and (ii) \textit{clusters}, which group together triples that share a contextual feature.

\paragraph{Strata.}
%We define strata based on entity labels, making the design entirely independent from \ac{NEL}. Facts are assigned to one of five strata that aim at uniform cardinality and semantic coherence:
We assign triples to one of 5 strata defined in a fully deterministic way based on their entity labels:
\begin{enumerate}
    \item \textbf{DDF} (Disease, Disorder, or Finding);%: disease, disorder, and finding-related entities;
    \item \textbf{Microbiome + Bacteria};
    \item \textbf{Human + Animal + Anatomical Location};
    \item \textbf{Chemical + Gene};
    \item \textbf{Drug + Dietary Supplement + Food + Biomedical Technique + Statistical Technique}.
\end{enumerate}
The definition of these strata took into account semantic coherence and uniform cardinality. 
By semantic coherence we mean grouping entity labels that tend to appear in similar contexts. For instance, \textit{Microbiome} and \textit{Bacteria} are grouped since bacteria compose the microbiome; \textit{Human}, \textit{Animal}, and \textit{Anatomical Location} are placed in the same stratum since they all describe biological subjects and their body parts; \textit{Chemical} and \textit{Gene} are aggregated since chemicals can be considered genes and viceversa depending on the context. 
Other groupings would have been possible and might yield to a better semantic coherence, but we also needed strata to contain a reasonably balanced number of triples. Therefore, the chosen stratification design represents the best trade-off we could obtain between semantic coherence and balanced stratum size, ensuring that no single stratum excessively dominates the estimation.
%Each stratum $h$ contains a subset $F_{h} \subset F$ of facts with specific entity types, but the sampling design does not use any \ac{NEL}-related features (e.g., concept frequency).

\paragraph{Clusters.}
Within each stratum, we define cluster groups as all triples whose mention text spans normalize to the same surface form.
Let $ts_f$ be the mention text span of a triple $t$, and let $\sigma_j$ indicate the normalized surface form associated with cluster $j$. Then, we can define a generic cluster $j$ as:  
\begin{align}
    \text{Cluster}_j &= \{ t \in T \;|\; \text{normalize}(ts_f) = \sigma_j \}%, \\
\end{align}

\begin{comment}
\begin{align}
    \text{Cluster}_j &= \{ t \in T \;|\; \text{normalize}(\text{textspan}_t) = \text{surface}_j \}%, \\
    %M_j &= |\text{Cluster}_j|,
\end{align}
\end{comment}

where normalization lowercases the text span and removes extra whitespaces. 
This design makes annotators reviewing multiple occurrences of the same surface form in sequence, thus reducing context switches and improving annotation efficiency.
Moreover, combined with the way of annotating incorrect links described in Section \ref{sec:accuracyEstimationForFactCollections}, this clustering approach enables downstream corrections at the cluster level, for example by correcting all at once mentions in the same cluster that were mapped to an over-general or wrongly assigned concept.
%systematic cluster-level corrections, e.g., by revising all mentions in the same cluster that were mapped to an over-general or wrongly assigned concept.

Table \ref{tab:samplingResults} shows the number of triples and unique clusters for each stratum. The higher weight of the \textit{DDF} stratum reflects the fact that this label actually groups three categories (diseases, disorders, and findings), which occur much more frequently than other entity types in the PubMed documents composing the \textsc{GutBrainIE} collection.

\paragraph{STWCS procedure.}
Sampling is done iteratively and consists of three steps:
\begin{enumerate}
    \item \textbf{Stratum selection}: sample a stratum $h$ with \ac{PPS-WR}, where size is the total number of triples $M[h]$ in the stratum.
    \item \textbf{Cluster selection}: within the selected stratum, sample a cluster $j$ with \ac{PPS-WOR}, considering as size the number of triple in the cluster ($|\text{Cluster}_j|$).
    \item \textbf{Triples annotation}: within the selected cluster, annotate up to $m$ triples (or all triples if the cluster is smaller).
\end{enumerate}
In step (1), only strata with unsampled clusters are considered. In step (3) we fix $m=5$, proven by \cite{gao2019efficient} to be a near-optimal choice that minimizes evaluation cost for \ac{TWCS} across different \acp{KG} while preserving its efficiency gains over \ac{SRS}.
After each iteration, we update the accuracy estimates and associated \ac{CI}, and we continue sampling and annotating until the target \ac{MoE} is reached. 
It's worth noting that both stratification and clustering are defined independently of \ac{NEL} outputs. %, avoiding circularity between what is being evaluated and how the evaluation sample is constructed.
This is crucial since we are evaluating \ac{NEL} itself and, therefore, cannot rely on its (a priori unknown) quality in our sampling strategy. This is reinforced by the fact that, as discussed in Section \ref{sec:nel_gutbrainie}, \ac{NEL} annotations are produced automatically. Our design choices avoid circularity between what is being evaluated and how the evaluation sample is constructed.

\subsection{Estimator and Confidence Interval}
We first recall the estimator used by \ac{TWCS} within a single (unstratified) set of triples $T$.
Let $n$ be the number of sampled clusters and, for the $k$-th sampled cluster, let $\mu_{I,k}$ indicate the mean accuracy of the (at most $m$) triples annotated in that cluster. The \acs{TWCS} estimator of the overall accuracy in this setting is given by:
\begin{equation}
    \hat{\mu}_{w,m} = \frac{1}{n} \sum_{k=1}^{n} \mu_{I,k}
\end{equation}
which is proven to be an unbiased estimator of the true accuracy under the \ac{TWCS} sampling design \cite{gao2019efficient}.

\ac{STWCS} extends \ac{TWCS} by applying it independently within each stratum and then aggregating the resulting estimates. 
Let $M[h]$ be the total number of triples in stratum $h$, $M = \sum_h M[h]$ the total number of triples, and $W_h = M[h]/M$ the weight of stratum $h$. 
Let $\hat{\mu}_{w,m,h}$ be the \ac{TWCS} estimator computed within stratum $h$. The \acs{STWCS} estimator of the global accuracy is then:
\begin{equation}
\label{eq:STWCS_estimator}
    \hat{\mu}_{ss} = \sum_h W_h \cdot \hat{\mu}_{w,m,h}
\end{equation}

and the $(1-\alpha)$ \ac{CI} for $\hat{\mu}_{ss}$ can be written as:
\[
\hat{\mu}_{ss} \pm z_{\alpha/2} \sqrt{\sum_h W_h^2 \cdot \mathrm{Var}(\hat{\mu}_{w,m,h})},
\]
where the variance term is estimated from the sampled clusters in each stratum. As shown in \cite{gao2019efficient}, the \ac{STWCS} estimator in Eq. \ref{eq:STWCS_estimator} is unbiased for the true corpus-level accuracy $\mu(T)$ under the sampling design described in Section \ref{sec:samplingDesign}.

\begin{comment}
Let $M[h]$ be the total number of facts in stratum $h$, $M = \sum_h M[h]$ the total number of facts, and $W_h = M[h]/M$ the weight of stratum $h$. Then, the \ac{STWCS} estimator of overall accuracy is: 
\begin{equation}
\label{eq:STWCS_estimator}
    \hat{\mu}_{ss} = \sum_h W_h \cdot \hat{\mu}_{w,m,h},
\end{equation}
where $\hat{\mu}_{w,m,h}$ is the within-stratum estimator produced by \acs{TWCS}, combining cluster-level accuracies with their sampling weights \cite{gao2019efficient}. Let $\mu_{Ik}$ be the mean accuracy of the sampled triples (at most $m$) in the $k$-th sampled cluster of a generic fixed stratum. The associated \acs{TWCS} estimator is:
\begin{equation}
    \hat{\mu}_{w,m}=\frac{1}{n}\sum_{k=1}^n \mu_{I,k} 
\end{equation}

The $(1-\alpha)$ \ac{CI} for $\hat{\mu}_{ss}$ can be written as
\[
\hat{\mu}_{ss} \pm z_{\alpha/2} \sqrt{\sum_h W_h^2 \cdot \mathrm{Var}(\hat{\mu}_{w,m,h})},
\]
%where $z_{\alpha/2}$ is the standard normal quantile and the variance term is estimated from the sampled clusters in each stratum. 
\cite{gao2019efficient} proves that the \ac{STWCS} estimator in Eq. \ref{eq:STWCS_estimator} is unbiased for $\mu(F)$ under the sampling design described in Section \ref{sec:samplingDesign}.
\end{comment}

\subsection{Cost Model and Context-Switching}

To assess efficiency, we introduce a simple cost model that captures the impact of context switches on annotation time. We define a \textit{context switch} as any transition between two consecutively annotated triples belonging to different surface-form clusters. Intuitively, switching clusters requires the annotator to adapt to a new textual and conceptual context, increasing annotation time.

Let $n_t$ be the total number of annotated triples in a sample $T_S$, and let $n_{sw}$ be the number of context switches occurred while annotating $T_S$.
We model the cost of annotating $T_S$ as
\begin{equation}
    \text{cost}(T_S) = c_t \cdot (n_t - n_{\text{sw}}) + c_s \cdot n_{\text{sw}}
\end{equation}
where $c_t$ is the per-triple cost without context switch and $c_s$ is the per-triple cost when a context switch occurs.

In our analysis, we measure cost in terms of wall-clock time. Let $t_{\text{base}}$ denote the mean annotation time without context switch and $\Delta t_{\text{switch}}$ the additional time introduced by switching context. The associated time-cost model can be written as:
\begin{equation}
    \text{time}(T_S) = n_t \cdot t_{\text{base}} + n_{\text{sw}} \cdot \Delta t_{\text{switch}}
\end{equation}
%Given time statistics measured on the actual annotation campaign, this model allows us to compare the expected total annotation time of STWCS against a baseline such as SRS, for a fixed number of annotated facts.
In the following, we will discuss wall-clock time results measured on the actual annotation campaign.

\subsection{Annotation Interface}
\label{sec:Annotation Interface}
To collect correctness annotations on sampled triples, we implemented a custom web application using \textit{streamlit}.\footnote{https://streamlit.io/}
This application supports the annotation workflow described in Section \ref{sec:accuracyEstimationForFactCollections}, allowing annotators to decide, for each triple, whether the link between mention and concept is correct, incorrect because mapped to an overly general concept, or incorrect because the linked concept is wrong.
The application also records the wall-clock time spent in annotating each triple, which is later used to compute values for the cost model illustrated in Section \ref{sec:samplingDesign}. Moreover, our application supports collaborative work by allowing annotators to export their annotations as a JSON file that can be used to resume their work or extend the one done by other experts.

\paragraph{Input Format.}
The input expected by our web application is a JSONL file containing the triples to be annotated and, optionally, a JSON file with annotations from a previous sessions (either by the same or another annotator).
To simplify the setup and avoid having to implement real-time iterative sampling, we generated a static sample batch before starting the actual annotation process.
To do that, we simulated an annotation round by sampling triples with our \ac{STWCS} design (see Section \ref{sec:samplingDesign}), fixing the random seed to 42 for reproducibility.
To ensure that the resulting batch was large enough to reach the target \ac{MoE}, we assigned synthetic labels for correctness/incorrectness under a worst-case assumption for convergence, i.e., simulating an accuracy around 0.5 by alternating triples as correct and incorrect. This procedure created a static batch comprising 7,517 triples.
In the produced JSONL file, each line corresponds to a single triple, and lines are ordered according to the \ac{STWCS} sampling order. The annotation interface presents triples in the order they are reported in this file, thus preserving the strata and clusters structure imposed by \ac{STWCS}, which aims to minimize context switches.

\paragraph{\ac{MoE} Convergence Monitoring.}
To notify annotators when the number of reviewed triples was sufficient to reach $\text{MoE} \leq \varepsilon=0.05$, we implemented a separate monitoring script in Python.
This script is launched before starting an annotation session and then keeps running on the background. 
Every time the annotator finishes evaluating all triples belonging to a sampled cluster, the script automatically recomputes the estimated accuracy and associated \ac{MoE}. If the computed \ac{MoE} goes below the target threshold $\varepsilon$, the script triggers a desktop notification (using the \textit{plyer} Python package\footnote{https://github.com/kivy/plyer}), telling the annotator that convergence has been reached and further annotations are not needed for the global accuracy estimate.

\begin{figure}[htb]
    \centering
    \includegraphics[width=\linewidth]{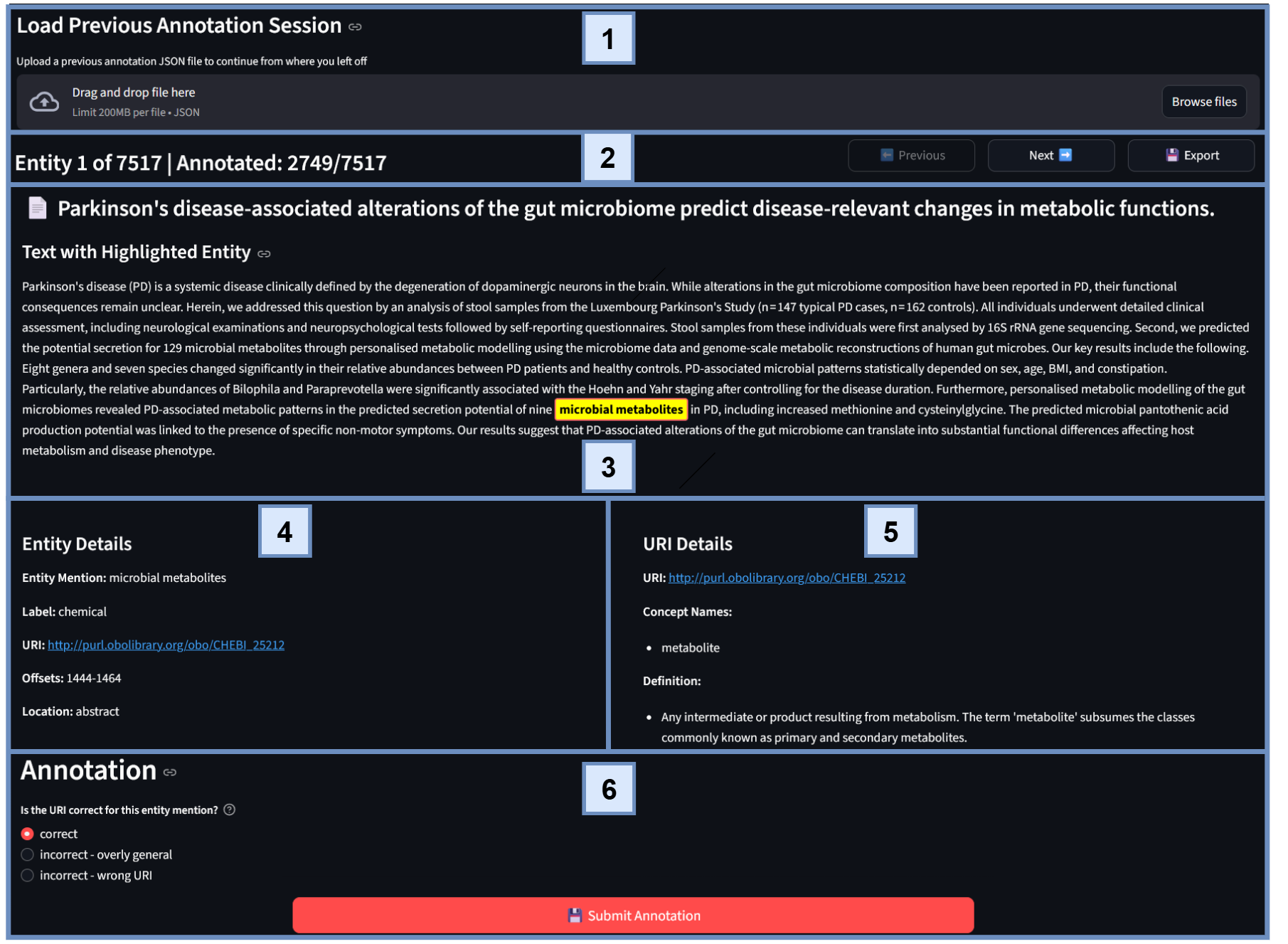}
    \caption{Annotation interface used for the revision of \ac{NEL} triples.}
    \label{fig:annotationInterface}
\end{figure}

\paragraph{User Interface.}
The annotation interface is shown in Figure \ref{fig:annotationInterface}. Panel (1) allows annotators to load a saved JSON file containing previous annotations, enabling them to resume work from another annotator or from an earlier annotation session.
Panel (2) summarizes progress, displaying the index of the current triple, the number of annotated triples over the total, and providing navigation controls. By double-clicking on the ``\textit{Next}'' arrow, annotators can open a dialogue box and directly jump to a specific triple index, which is useful when continuing previous sessions. This panel also includes an ``\textit{Export}'' button that allows annotators to save all annotations produced so far to a JSON file with a custom name and to a custom location. Notice that the application still automatically saves annotations to a JSON file in a local cache folder every time the user moves to a different triple.  
Panel (3) shows the full abstract text with the current mention highlighted, giving annotators all necessary local context.
Panels (4) and (5) display details about the entity and the linked concept. Specifically, (4) includes the mention text, label, location (title/abstract) and offsets in the location (in terms of number of characters, including whitespaces), while (5) reports the \ac{URI} along with associated names and definitions retrieved from the reference resource. Finally, panel (6) contains the annotation controls and a button to submit the rating and proceed to the next triple. 

\section{Results}
\label{sec:Results}

\subsection{Annotation Campaign and Sampling Outcome}

The expert folds of \textsc{GutBrainIE} contain a total of 11,184 \ac{NEL} triples organized into 4,116 global surface-form clusters across the five strata described in Section \ref{sec:samplingDesign}. Using the \ac{STWCS} design, we iteratively sampled and annotated triples until reaching a target \ac{MoE} ($\varepsilon$) of 0.05 for the overall accuracy estimate.

Convergence was obtained after annotating 2,749 triples (24.6\% of all triples), belonging to 1,044 unique clusters. Across subsequent annotations, we observed 1,050 context switches and 1,698 no-switch transitions (i.e., transitions within the same cluster).

The resulting overall \ac{NEL} accuracy estimate is
\[
\hat{\mu}_{ss} = 0.915 \pm 0.0473,
\]
corresponding to a $(1-\alpha)$ \ac{CI} of $[0.868, 0.963]$. 
The resulting \ac{MoE} is $0.047$, thus below the target threshold $\varepsilon=0.05$. 
%This estimate summarizes the accuracy of all NEL facts in the expert-curated folds of \textsc{GutBrainIE}.

\subsection{Accuracy Estimates by Stratum}

\ac{STWCS} also produces per-stratum accuracy estimates. The number of annotated clusters and triples for each stratum is reported in Table \ref{tab:samplingResults}, while the estimated accuracy $\hat{\mu}$, the corresponding \ac{MoE}, and the resulting $(1-\alpha)$ \ac{CI} for each stratum are reported in Table \ref{tab:accuracyEstimates}.

\begin{table}[htb]
\caption{Distribution of weights, clusters, and triples across strata. For each stratum, the \textit{Clusters} and \textit{Triples} columns report the number of sampled (i.e., annotated) items over the total available.}
\label{tab:samplingResults}
\resizebox{\textwidth}{!}{%
\begin{tabular}{|l|c|c|c|}
\hline
Stratum                                                             & Weight & \begin{tabular}[c]{@{}c@{}}Clusters\\ (Annotated / Total)\end{tabular} & \begin{tabular}[c]{@{}c@{}}Triples\\ (Annotated / Total)\end{tabular} \\ \hline
DDF                                                                 & 0.3670 & 366 / 1,321                                                            & 1,013 / 4,105                                                       \\ \hline
Microbiome + Bacteria                                               & 0.1815 & 197 / 514                                                              & 513 / 2,030                                                         \\ \hline
Human + Animal + Anatomical Location                                & 0.1858 & 209 / 665                                                              & 559 / 2,078                                                         \\ \hline
Chemical + Gene                                                     & 0.1330 & 147 / 802                                                              & 381 / 1,487                                                         \\ \hline
Drug + Dietary Supplement + Food + Biomedical/Statistical Technique & 0.1327 & 128 / 814                                                              & 283 / 1,484                                                         \\ \hline
\end{tabular}%
}
\end{table}

\begin{table}[htb]
\caption{Strata-level \ac{NEL} accuracy estimates, reporting for each stratum the estimated $\hat{\mu}$ and the corresponding margin of error (\ac{MoE}).}
\label{tab:accuracyEstimates}
%\resizebox{\textwidth}{!}{%
\begin{tabular}{|l|c|c|}
\hline
Stratum & $\hat{\mu}$ & \ac{MoE} \\ \hline
DDF & 0.883 & 0.093 \\ \hline
Microbiome + Bacteria & 0.979 & 0.108 \\ \hline
Human + Animal + Anatomical Location & 0.976 & 0.060 \\ \hline
Chemical + Gene & 0.851 & 0.087 \\ \hline
Drug + Dietary Supplement + Food + Biomedical/Statistical Technique & 0.898 & 0.154 \\ \hline
\end{tabular}%
%}
\end{table}

\begin{comment}
\begin{table}[htb]
\centering
\caption{\ac{NEL} accuracy estimates by label-based stratum. \acp{CI} are $(1-\alpha)$ intervals obtained under the \ac{STWCS} estimator.}
\label{tab:strata-results}
\resizebox{\textwidth}{!}{%
\begin{tabular}{|l|c|c|c|c|c|c|c|}
\hline
Stratum & \#facts & \#clusters & Weight & Sampled clusters & $\hat{\mu}$ & CI width & $\hat{\mu}$ bounds \\ \hline
DDF & 4,105 & 1,321 & 0.3670 & 364 & 0.883 & 0.093 & [0.790, 0.976] \\ \hline
Microbiome + Bacteria & 2,030 & 514 & 0.1815 & 196 & 0.979 & 0.108 & [0.871, 1.087] \\ \hline
Human + Animal + Anatomical Location & 2,078 & 665 & 0.1858 & 209 & 0.976 & 0.060 & [0.916, 1.036] \\ \hline
Chemical + Gene & 1,487 & 802 & 0.1330 & 147 & 0.851 & 0.087 & [0.764, 0.938] \\ \hline
Drug + Dietary Supplement + Food + Biomedical/Statistical Technique & 1,484 & 814 & 0.1327 & 128 & 0.898 & 0.154 & [0.744, 1.052] \\ \hline
\end{tabular}%
}
\end{table}
\end{comment}

While strata-level intervals are wider than the overall \ac{CI} (as expected given the reduced number of triples annotated per group), they provide useful insights into how \ac{NEL} accuracy varies across different groups of entity types. Achieving narrower strata-level \acp{CI} would require additional annotations, which our framework supports by continuing sampling in specific strata.
 
Overall, if we focus on the lower bounds of the strata-level \acp{CI} ($\hat{\mu}-\text{MoE}$) as conservative estimates of accuracy, we observe for all strata an accuracy estimate of at least $0.74$, indicating satisfactory \ac{NEL} quality across all considered entity groups.
In particular, \textit{Microbiome + Bacteria} and \textit{Human + Animal + Anatomical Location} strata obtain lower bounds above $0.85$, highlighting high \ac{NEL} reliability for these entity labels.
These strata-level estimates can be leveraged as guidance for prioritizing downstream targeted curation on strata showing lower accuracy.

\subsection{Efficiency Analysis vs Simple Random Sampling}

To quantify the efficiency gains coming from our \ac{STWCS} sampling design, we analyze annotation times and compare them to a simulated \ac{SRS} baseline.
About the latter, since re-running the entire annotation campaign under \ac{SRS} would be unnecessarily costly, we estimate it via simulation. Specifically, we keep fixed the same 2,749 triples annotated under \ac{STWCS} and simulate what would happen if they were drawn and annotated with \ac{SRS} instead (i.e., with uniform random probability). This produces a random permutation of the triples, which we use to count the number of transitions with and without context switches. We repeat this process 1,000 times and then bootstrap the resulting transition counts to reduce the impact of rare (i.e., outlier) \ac{SRS} permutations.  

The total time spent by experts to annotate the 2,749 sampled triples under \ac{STWCS} was 797 minutes (13 hours and 17 minutes). From the logs collected during the annotation campaign, we estimate:
\begin{itemize}
    \item mean annotation time without context switch ($t_{\text{base}}$) of $0.22$ minutes (12.97 seconds);
    \item mean annotation time with context switch ($t_{\text{base}} + \Delta t_{\text{switch}}$) of $0.41$ minutes (24.57 seconds),
\end{itemize}
From that, we can derive that the additional cost per context switch ($\Delta t_{\text{switch}}$) is 0.19 minutes (\text{11.59 seconds}). Therefore, switching context introduces a slowdown factor of $0.41 / 0.22 \approx 1.89$.

%We then simulated an evaluation campaign on the same set of 2,749 facts, but assuming \ac{SRS} without any clustering or stratification. We conducted 1,000 simulations, and in each we randomly permuted the facts, counted the number of context switches between consecutive annotations, 
To provide a realistic estimate of the annotation cost that \ac{SRS} would have, we estimate the total annotation time under \ac{SRS} using the same $t_{\text{base}}$ and $\Delta t_{\text{switch}}$ measured above. 
We bootstrapped the distribution resulting from 1,000 \ac{SRS} simulations and obtained a mean number of context switches of $2,745$, a mean number of no-switch transitions of $2.84$, and a mean total annotation time of 1,124.57 minutes (18 hours and 44 minutes).

Compared to this \ac{SRS} baseline, \ac{STWCS} saves approximately 327.6 minutes (5 hours and 28 minutes) of expert time at fixed sample size. The corresponding efficiency ratio is
$\frac{\text{time}_\text{STWCS}}{\text{time}_\text{SRS}} \approx 0.71$,
indicating that our sampling framework achieves the same statistical precision with about 29\% less annotation time.
This demonstrates that applying \ac{STWCS} to \ac{NEL} accuracy estimation gives efficiency gains over \ac{SRS} that are aligned to those reported by \cite{gao2019efficient,marchesin2024utility} for \ac{KG} accuracy estimation, even though our setting focuses on \ac{NEL} triples rather than \ac{KG} triples.

\section{Conclusion and Future Directions}
\label{sec:Conclusion}
We presented a sampling-based framework for estimating \ac{NEL} accuracy in the \textsc{GutBrainIE} collection under statistical guarantees and limited annotation budget. By framing \ac{NEL} accuracy estimation as a constrained optimization problem and adapting the \ac{STWCS} design to \ac{NEL} triples, we were able to estimate a corpus-level accuracy of $0.915 \pm 0.0473$ by annotating less than 25\% of all triples and reducing the total annotation time by about 29\% compared to a simulated \ac{SRS} baseline, demonstrating the effectiveness of our sampling design focused on minimizing context switches. Moreover, we were able to obtain strata-level accuracy estimates representing \ac{NEL} accuracy across groups of entity types.
\begin{comment}
\begin{itemize}
    \item obtain a corpus-level \ac{NEL} accuracy estimate of $0.915 \pm 0.0473$ by annotating less than 25\% of all facts;
    \item derive strata-level accuracy estimates representing performance differences across groups of entity types;
    \item reduce total annotation time by about 29\% compared to a simulated \ac{SRS} baseline, leveraging our \ac{STWCS}-based clustering strategy that minimizes context switches.
\end{itemize}
\end{comment}

Our design uses only \ac{NEL}-independent features (entity labels and surface forms) to define strata and clusters, avoiding circularity between the evaluation target and the sampling approach. 
Moreover, it gives the possibility to continue sampling in specific strata to tighten local \acp{CI} (i.e., strata-level \acp{CI}) when needed.
Finally, given that the proposed framework is not specific to \textsc{GutBrainIE}, it can be applied to other \ac{NEL} benchmarks and \ac{IE} pipelines that require scalable and statistically robust accuracy assessment under limited expert budgets.

Future work might employ multiple annotators per triple and aggregate their labels (e.g., via majority voting), providing more reliable annotations on the correctness of linked concepts. Moreover, stratification could incorporate downstream impact, for instance, by prioritizing concepts that participate in many relation instances or have high centrality in the induced \ac{KG}. 

\begin{acknowledgments}
  This project has received funding from the HEREDITARY Project, as part of the European Union's Horizon Europe research and innovation programme under grant agreement No GA 101137074.    
\end{acknowledgments}

%% The declaration on generative AI comes in effect
%% https://ceur-ws.org/GenAI/Policy.html
\section*{Declaration on Generative AI}
During the preparation of this work, the author used GPT-4o and Grammarly in order to: Grammar and spelling check. After using these tools, the author reviewed and edited the content as needed and takes full responsibility for the publication’s content. 
  
%%
%% Define the bibliography file to be used
\bibliography{sample-ceur}

\end{document}